\documentclass[letterpaper, 10 pt, conference]{ieeeconf}  
\usepackage{graphicx}
\usepackage{placeins}
\usepackage{algorithm} 
\usepackage{algorithmicx} 
\usepackage{algpseudocode}
\usepackage{xcolor}
\usepackage{afterpage}
\usepackage{hyperref}
\usepackage{booktabs}

\usepackage{amsmath}  

\IEEEoverridecommandlockouts                              

\title{\LARGE \bf
ORB: Operating Room Bot, Automating Operating Room Logistics through Mobile Manipulation$^*$
}

\author{
    Jinkai Qiu$^{1,\dagger}$, 
    Yungjun Kim$^{1,\dagger}$, 
    Gaurav Sethia$^{1,\dagger}$, 
    Tanmay Agarwal$^{1,\dagger}$,
    Siddharth Ghodasara$^{1,\dagger}$, \\ 
    Zackory Erickson$^{1}$, 
    Jeffrey Ichnowski$^{1}$
    \\ 
\thanks{$^*$This work has been accepted to IEEE CASE 2025}
\thanks{$\dagger$ These authors contributed equally to this work.}
\thanks{$^{1}$Robotics Institute, Carnegie Mellon University, emails: \texttt{\{jinkaiq, yungjunk, gsethia, tanmaya, sghodasa\}\allowbreak@andrew.cmu.edu}, 
\texttt{\{zackory, jeffi\}@cmu.edu}}
}

\usepackage{afterpage}

\begin{document}

\maketitle
\thispagestyle{empty}
\pagestyle{empty}

\begin{abstract}
Efficiently delivering items to an ongoing surgery in a hospital operating room can be a matter of life or death. In modern hospital settings, delivery robots have successfully transported bulk items between rooms and floors. However, automating item-level operating room logistics presents unique challenges in perception, efficiency, and maintaining sterility. We propose the Operating Room Bot (ORB), a robot framework to automate logistics tasks in hospital operating rooms (OR). ORB leverages a robust, hierarchical behavior tree (BT) architecture to integrate diverse functionalities of object recognition, scene interpretation, and GPU-accelerated motion planning. The contributions of this paper include: (1) a modular software architecture facilitating robust mobile manipulation through behavior trees; (2) a novel real-time object recognition pipeline integrating YOLOv7, Segment Anything Model 2 (SAM2), and Grounded DINO; (3) the adaptation of the cuRobo parallelized trajectory optimization framework to real-time, collision-free mobile manipulation; and (4) empirical validation demonstrating an 80\% success rate in OR supply retrieval and a 96\% success rate in restocking operations. These contributions establish ORB as a reliable and adaptable system for autonomous OR logistics. Videos are available at https://www.youtube.com/watch?v=QTpJp1fD74c
\end{abstract}

\section{Introduction}

Modern OR facilities face escalating logistical challenges due to increased patient volume, workforce shortages, and stringent sterility requirements. Precise and timely OR supply retrieval and restocking operations are crucial for optimal surgical outcomes, yet reliance on human personnel introduces inefficiencies and contamination risks. Current automated logistics solutions, such as pneumatic tube systems and conveyor-based dispensing, offer efficiency but lack the flexibility required in dynamic OR environments, entail substantial upfront investments, and handle primarily fixed-location tasks.

Emerging autonomous mobile robots (AMRs), such as Aethon Tug, have improved hospital logistics by automating large-scale supply transport. However, these platforms typically lack precise manipulation capabilities essential for handling smaller, sterile OR supplies. Mobile manipulators like Diligent Robotics Moxi~\cite{moxi_web} offer promising dexterous manipulation capabilities but prioritize patient interaction and general-purpose tasks over rigorous, repetitive logistical workflows.

\begin{figure}[t]
  \centering
    \includegraphics[width=\columnwidth]{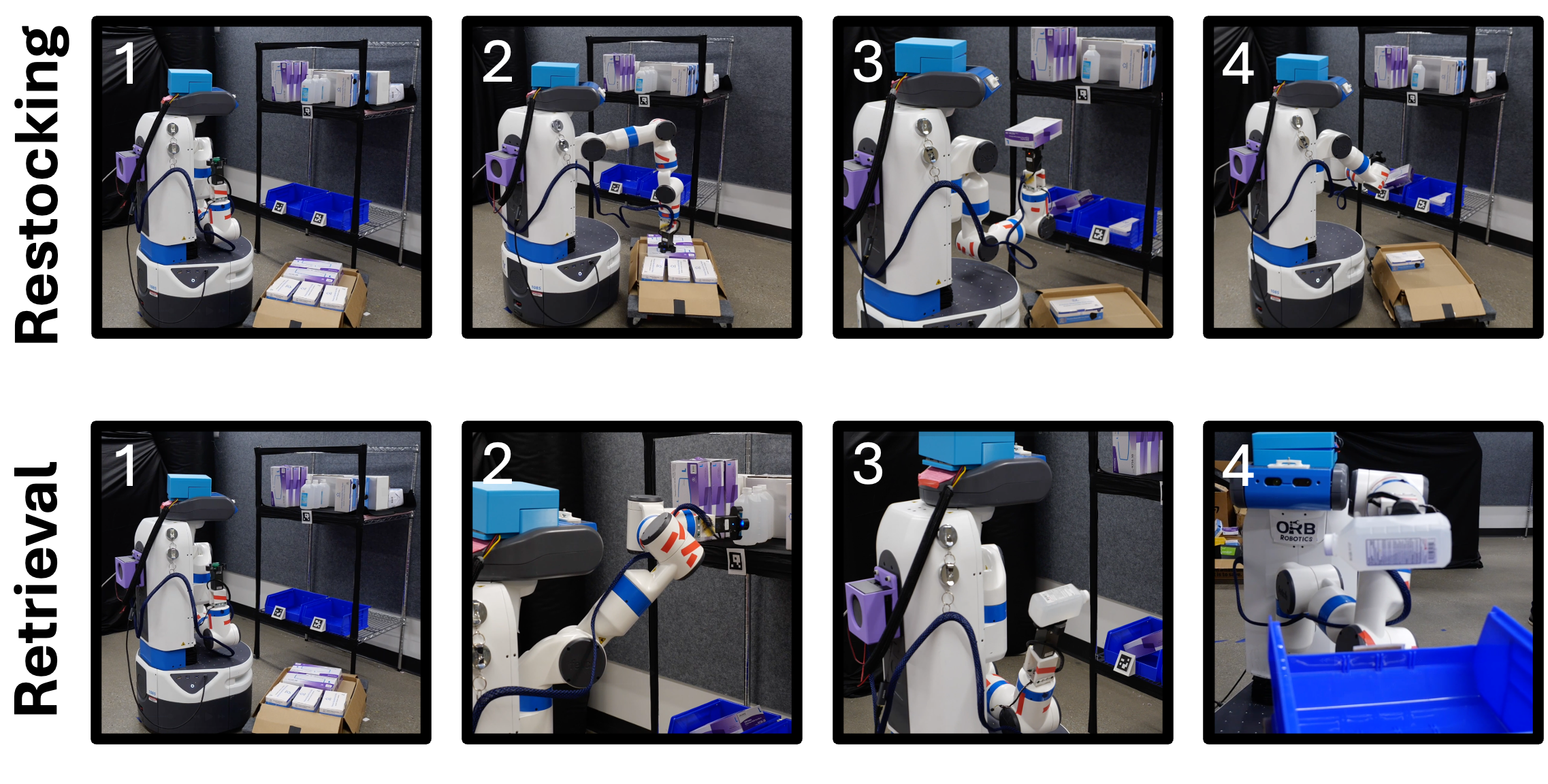}
  \caption{Operating Room Bot (ORB) performing two tasks in an operating room logistics environment. The top row demonstrates \textbf{RESTOCKING}, where ORB grabs OR items from a cart and places them onto storage shelves. The bottom row illustrates \textbf{RETRIEVAL}, where ORB picks up specific items from the shelves and delivers the item to a point of request.}
  \label{experiments}
  \vspace{-16pt}
\end{figure}

We propose Operating Room Bot (ORB) to address these limitations by introducing an autonomous mobile manipulation system tailored for OR logistics, including retrieval and restocking of OR supplies, as shown by Figure \ref{experiments}. Built on the Fetch mobile manipulator with a seven-degree-of-freedom (7-DOF) arm, precision suction gripper, and multimodal sensors, ORB employs a modular ROS2-based software architecture.
The behavior-tree-based decision-making framework ensures robust execution of tasks and adaptability to environmental uncertainties.

This paper makes the following primary contributions: 
\begin{itemize}
\item A modular behavior-tree-based architecture that integrates traditional engineering pipelines with learned models to create a robust and efficient mobile manipulation system.
\item A real-time scene and object understanding pipeline integrating state-of-the-art deep learning methods for reliable object recognition and segmentation.
\item Integration of real-time, GPU-accelerated motion planning using cuRobo~\cite{curobo} to mobile manipulation, tailored for precise and safe mobile manipulation tasks in constrained hospital environments. 
\item Real-world experiments suggesting ORB's efficacy and suitability for practical hospital deployments.
\end{itemize}

\begin{figure*}[t]
  \centering
    \includegraphics[width=\textwidth]{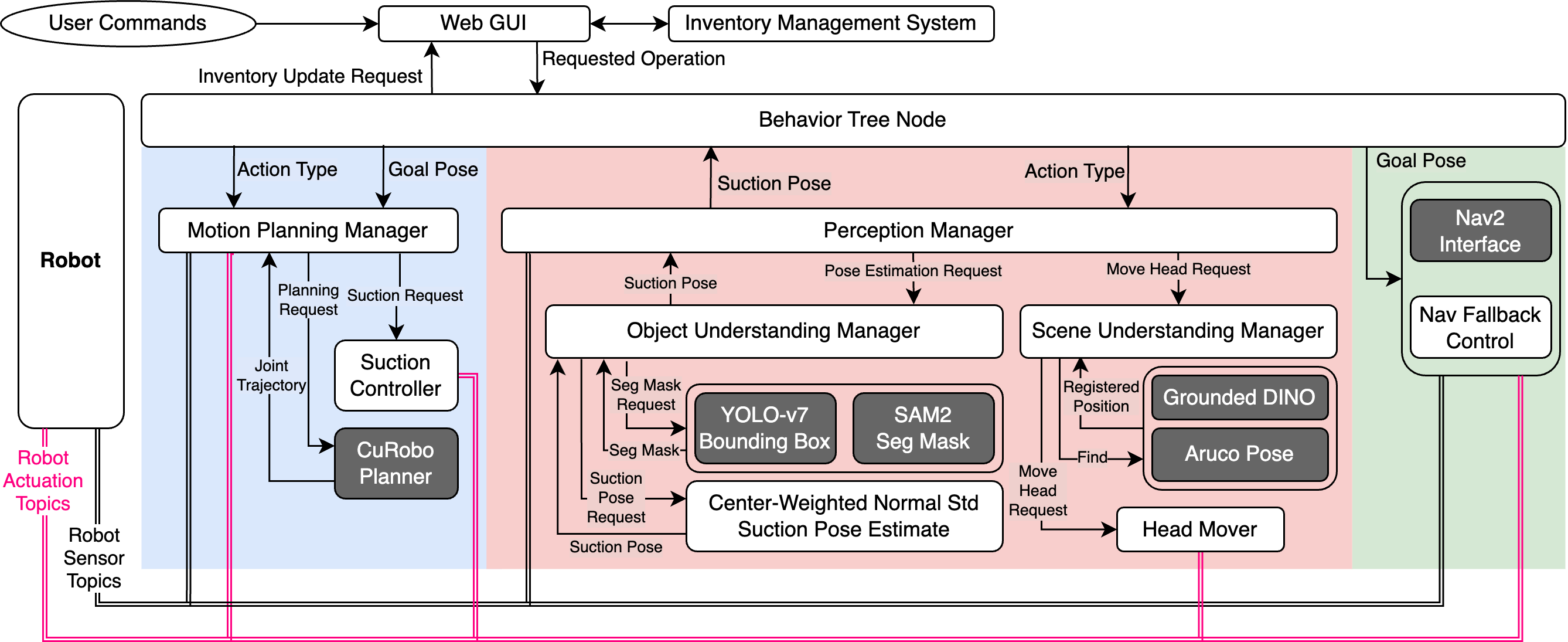}
  \caption{Software Architecture of ORB. \textit{\textcolor{blue}{Blue}} indicates manipulation subsystem. \textit{\textcolor{red}{Red}} indicates perception subsystem; \textit{\textcolor{green}{Green}} indicates navigation subsystem; \textit{Gray} boxes indicates adapted off-the-shelf components.}
  \label{system_overview}
  \vspace{-16pt}
\end{figure*}
\section{Related Work}

Mobile manipulation has been widely studied in robotics, integrating autonomous navigation with dexterous manipulation to perform complex tasks. Research in mobile manipulation spans system design, decision-making frameworks, perception techniques, and real-time motion planning, each contributing to improved efficiency and adaptability~\cite{thakar2022survey}. Nevertheless, gaps persist, particularly regarding the integration of adaptive perception methods, reliable manipulation techniques, and robust behavior execution in highly dynamic OR environments. ORB addresses these gaps by combining state-of-the-art perception modules, a reliable suction-based grasping strategy, and a flexible Behavior Tree framework, thereby adapting the robotic system for adaptability, reliability, and efficiency in realistic OR logistics tasks.


\subsection{Mobile Manipulation in Healthcare and Logistics}
Mobile manipulators have been increasingly deployed in healthcare settings for supply retrieval, patient assistance, and logistics automation. Robots such as Moxi \cite{moxi_web} focus on human-robot interaction, assisting hospital staff and elderly individuals with routine tasks~\cite{technologies9010008}. Autonomous mobile robots like Aethon Tug~\cite{tug} are used for hospital material transport but lack dexterous manipulation capabilities. Research in human-aware motion planning has introduced adaptive speed control and trajectory optimization to ensure safe interactions in dynamic hospital environments~\cite{Deepak03032016}. The integration of perception, planning, and control frameworks continues to expand the applicability of mobile manipulators in healthcare settings, improving efficiency and reducing operational burdens on human staff~\cite{health_mm}. In contrast to prior solutions, ORB focuses on requirements unique to the operating room setting.

\subsection{Decision-Making Frameworks for Mobile Manipulation}
Decision-making in mobile manipulation involves both high-level task planning and low-level motion execution. Traditional approaches rely on finite-state machines and rules-based control~\cite{506501}, which provide structured task execution but lack adaptability to dynamic changes. More recent strategies incorporate behavior trees (BT) \cite{Colledanchise_2018} and task and motion planning (TAMP) methods to improve modularity and real-time responsiveness~\cite{Zhao_2024}. The increasing complexity of tasks has also led to the use of large language models for general robot control~\cite{black2024pi0visionlanguageactionflowmodel}. As the OR setting has unique robustness and explainability requirements, ORB employs BT.

\subsection{Perception for Mobile Manipulation}
Perception is essential for enabling robots to understand their environment and interact with objects accurately. Mobile manipulators rely on RGB-D cameras, LiDAR, and depth sensors for real-time scene reconstruction and object recognition. Deep-learning models, including convolutional neural networks and transformer-based architectures, have significantly improved object detection and segmentation performance, allowing robots to handle diverse and unstructured environments~\cite{Bavle_2023}. For prediction of suction pose, both SuctionNet~\cite{Cao_2021} and DexNet~\cite{mahler2019learning} showed remarkable performance in their benchmark data sets, demonstrating their potential for reliable manipulation of various medical supplies in highly complex OR settings.  

\subsection{Real-Time Motion Planning for Mobile Manipulation}
Motion planning techniques for mobile manipulators focus on generating collision-free trajectories while optimizing task efficiency. Sampling-based algorithms such as rapidly-exploring random trees (RRT) and probabilistic roadmaps (PRM) are widely used for path planning in high-dimensional spaces~\cite{orthey2023samplingbasedmotionplanningcomparative}. Optimization-based approaches, including CHOMP, STOMP, and TrajOpt, enhance motion smoothness, predictability, and execution time by refining planned trajectories~\cite{Zhao_2024}. Recent advancements in GPU-accelerated trajectory optimization, such as cuRobo~\cite{curobo}, have improved real-time performance, enabling faster collision checking and dynamic re-planning. This project utilizes cuRobo to enable fast, predictable, and reliable planning of the manipulator, ensuring it meets the time-sensitive and high-reliability requirements of OR logistics.
\section{Method and System Architecture} \label{method}
The Operating Room Bot (ORB) system architecture (Figure~\ref{system_overview}) is designed for scalable mobile manipulation task orchestration. We designed ORB to restock regular items (boxes, bottles), which empirically includes the majority of inventory items at the OR, from a rolling cart and retrieve items from an inventory shelf. Built on the Fetch mobile manipulator ~\cite{wise_fetch_2016}, the ORB system integrates a differential-drive mobile base with a seven-degree-of-freedom (7-DOF) robotic arm installed on top of a torso prismatic joint, complemented by a suction gripper (CZ EVS01~\cite{chengzhou_vacuum_gripper}), external battery augmentation, and onboard Nvidia Jetson AGX Orin~\cite{nvidia_jetson_orin} (ORB Compute) computing unit for enhanced real-time computation. The ORB software stack, developed on ROS2, incorporates three main subsystems: perception, manipulation, and navigation, coordinated through a behavior-tree-based decision framework for robust task execution.

\subsection{Behavior Tree Architecture}
\begin{figure}[t]
  \centering
    \includegraphics[scale=0.4]{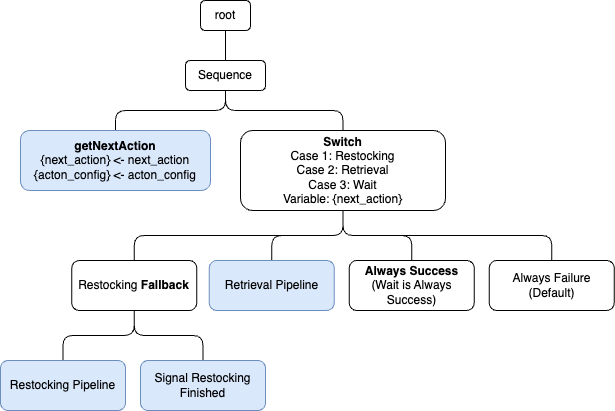}
  \caption{High-Level Task Selection logic for ORB. \textit{\color{blue}{Blue}} boxes indicate mid-level strategies of actions, \textit{white} boxes indicate behavior tree logic}
  \label{high_level_task}
\end{figure}

\begin{figure}[t]
  \centering
    \includegraphics[scale=0.4]{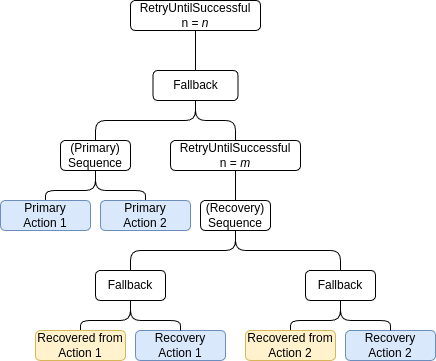}
  \caption{Low-level fallback framework. \textit{\textcolor{blue}{Blue}} boxes represent user-defined actions, \textit{\textcolor{brown}{yellow}} boxes represent conditions, and \textit{white} boxes represent behavior tree logic. \textit{n} and \textit{m} are tunable parameters that define the number of trials before giving up.}
  \label{low_level_fallback}
  \vspace{-16pt}
\end{figure}

ORB employs a hierarchical Behavior Tree (BT) implemented in ROS2 utilizing the \textit{BehaviorTree.CPP} library~\cite{BehaviorTreeCPP}. BT nodes interface with ROS2 through service calls, enabling asynchronous and modular action execution at a frequency of 5\,Hz. The BT design encompasses three hierarchical levels:
\begin{itemize}
    \item \textbf{High-Level Task Selection:} Manages task allocation (e.g., retrieval, restocking, idle state) using an externally managed job queue (Figure~\ref{high_level_task}).
    
    \item \textbf{Mid-Level Task Strategies:} Defines sequences of robot actions required for specific tasks.
    
    \item \textbf{Low-Level Action Recovery Mechanisms:} Implements fallback behaviors on failure conditions, ensuring resilience through recovery strategies such as navigation adjustments, replanning manipulation trajectories, and responding to unexpected object handling failures (Figure~\ref{low_level_fallback}).
\end{itemize}
This structured and modular approach allows ORB to handle task complexity and real-time contingencies effectively.

\subsection{User Interface}
ORB employs a graphic user interface (GUI) to initiate the High-Level Task Selection process, where the user could request item(s) to be retrieved or restocked. In the back-end, the robot will match item names to their storage locations and send for further processing. All tasks are queued in the order they were received, and retrieval tasks are prioritized over restocking due to their emergent nature. 

\subsection{Scene Understanding}
ORB's scene understanding pipeline is optimized for the rapid identification and localization of restocking objects within cluttered hospital supply environments. We utilize \textit{Grounding DINO Tiny}~\cite{liu2024grounding} for real-time object detection, configured with the prompt \textit{``cardboard box''} and a threshold of 0.55, to balance detection sensitivity and computational efficiency on ORB Compute. We observed that OR shelves are already equipped with one QR code per item bin for inventory counting and ordering. Thus, we decided to use ArUco markers~\cite{GARRIDOJURADO20142280}, which off-the-shelf allow precise bin identification and estimation of 6-DOF bin poses to ensure accurate restocking.

\subsection{Object Recognition and Suction Pose Generation}
ORB's object recognition pipeline integrates state-of-the-art deep learning approaches for accurate object segmentation and suction pose generation. First, a custom-trained YOLO-v7~\cite{wang2022yolov7trainablebagoffreebiessets} model generates bounding box proposals, refined subsequently by Segment Anything Model 2 (SAM-v2)~\cite{ravi2024sam2segmentimages} to achieve highly precise segmentation masks. This two-stage strategy effectively balances computational efficiency with segmentation accuracy on ORB Compute. This custom-trained YOLO-v7 model was developed using a dataset manually collected in our simulated OR setup. The dataset comprised 500 images across three classes of OR objects relevant to our tasks. During training, data augmentation methods such as changes in brightness and contrast, rotation, and the application of salt and pepper noise were employed to enhance the model's generalization capabilities and robustness to environmental variations.

For robust estimation of suction pose, we use a modified version of the Normal-std approach proposed by SuctionNet~\cite{Cao_2021}. Empirically, this method is robust for our application, while learned methods such as SuctionNet itself fail to generalize to novel scenes outside its trained distribution. Similar to SuctionNet's implementation of Normal-std, this method utilizes Open3D~\cite{zhou2018open3dmodernlibrary3d} to estimate the surface normals of a depth image, then applies a 2D convolution mask to the normal map to compute the standard deviation of surface normals, obtaining a normal standard deviation heatmap \( H \). We then specifically augment $H$ by integrating a computed centerness score, which prioritizes suction poses toward object centers to maximize stability during grasp operations:
\begin{equation}
    C(p) =  1 - \frac{d(p, c)}{d_{\max}},
    \label{eq:centerness}
\end{equation}
where \( c \) is the center of the object segmentation mask, \( d(p, c) \) represents the Euclidean distance from pixel \( p \) to the center \( c \), \( d_{\max} \) is the maximum distance within the segmentation mask, and \( \epsilon \) is a tunable parameter. 

Given the computed centerness score \( C(p) \), the pipeline normalizes it by its mean value \( \bar{C} \) and scale it by a tunable parameter \( \epsilon \). Similarly,  \( H(p) \) is normalized by its mean value \( \bar{H} \) to maintain consistent scaling. To incorporate object centerness into the heatmap, we compute the center-weighted heatmap as follows:  
\begin{equation}
    H''(p) = \left( \frac{H(p)}{\bar{H}} \right) \cdot \left( \frac{C(p)}{\bar{C}} \cdot \epsilon \right),
    \label{eq:final_heatmap}
\end{equation}  
where \( \bar{C} \) and \( \bar{H} \) represent the average centerness score and heatmap value over the object mask, respectively. This formulation ensures a consistent scaling while integrating centerness information into the heatmap.
From \( H'' \), we extract the top-\( K \) highest-scoring pixels \( P_K = \{ p_1, p_2, ..., p_K \} \), and compute the final suction pose \( \mathbf{n}^* \) as the average of the normals at these points.

\subsection{Real-time Motion Planning}
ORB leverages cuRobo~\cite{curobo}, a GPU accelerated optimization xbased motion planning framework, to enable fast and reliable generation of collision-free trajectories. This planner outperforms traditional sampling-based methods by providing smoother trajectories with minimal computational overhead.

To enhance safety and precision, we predefined task-specific motion primitives, such as linear end-effector movements. These primitives allow ORB to plan arm trajectories predictably and safely in constrained hospital environments, which is crucial for operations involving delicate OR supplies and tight spaces. The behaviour tree (BT) controls and invokes the motion planner with the correct primitive informing it of the kind of motion it needs to perform in the specific scenario, for instance, when the vision pipeline determines a grasp pose, the BT invokes the motion planner with the reach goal primitive, which is interpreted by the planner to plan a constraint free path to the pre grasp pose and then a linearly constraint path to the final grasp pose, further on a successful grasp BT invokes the planner again but with a reverse constraint where the planner plans the path in reverse ensuring a safe and known path which takes into account the picked up object. Similarly the BT is responsible to invoke the other primitives like homing and placing depending on the progress and state of the current task. 

\begin{figure}[t]
  \centering
    \includegraphics[width=\columnwidth]{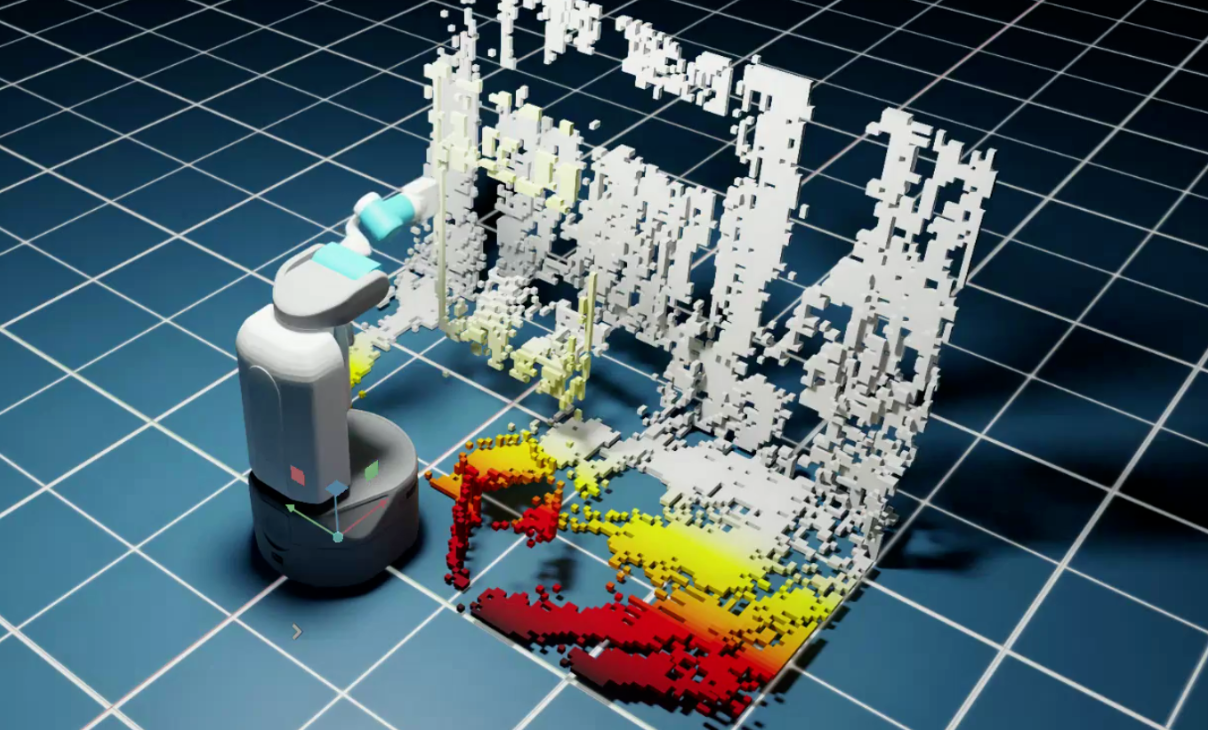}
  \caption{Visualization of NVBlox constructing a voxel-based representation of a real-world stocked OR supply shelf, as perceived by ORB’s RGB-D camera. The voxelized environment enables ORB to perform real-time collision avoidance and motion planning. Red indicates nearby obstacles, while white represents areas beyond reach.}
  \label{nvblox}
  \vspace{-16pt}
\end{figure}

The ORB system uses NVBlox~\cite{nvblox}, a voxel-based environment mapping framework, to efficiently generates truncated signed-distance fields (TSDFs) and Euclidean signed-distance fields (ESDFs) for real-time collision detection and avoidance, which continuously updates a voxel-based representation of the workspace during scene understanding phase (Figure \ref{nvblox}). For the current system we run the scene understanding phase whenever there is a change in the base position of the robot allowing it to tackle any obstacles in the scene before the arm starts moving.

To detect suction success/failure, the ORB system introduces a gripper feedback function into the motion planning system, enabling real-time trajectory preemption. This mechanism allows the robot to continuously assess the grasp status and abort the current motion plan upon a successful grasp or a detected anomaly, such as a failed suction or an unexpected object slip.

\subsection{Navigation} \label{nav}
ORB uses ROS2’s Nav2 stack~\cite{macenski2020marathon2} with Adaptive Monte Carlo Localization (AMCL) and onboard LiDAR for robust 2D navigation. While Nav2 offers reliable baseline performance, minor positional errors of around 10

 cm were frequently observed. Since the robot operates in a controlled indoor environment, external markers like QR codes and ArUco markers~\cite{GARRIDOJURADO20142280} are commonly available. To mitigate navigation inaccuracies, we implemented a reactive correction strategy that leverages ArUco detections to dynamically adjust the robot’s position using velocity commands when Nav2 fails to reach the goal. This recovery mechanism improves manipulation reliability by maintaining consistent alignment within the robot’s workspace constraints.

This architecture, combining robust navigation, high-precision perception, optimized motion planning, and sophisticated behavior orchestration, positions ORB as a practical solution capable of improving operational efficiency and reliability in hospital logistics environments.

\section{Experiments}
We conducted experiments in an environment that simulated a hospital supply room, with three designated items for retrieval and restocking. Our evaluations included both subsystem assessments (navigation, constrained motion planning, suction pose estimation, and scene understanding), and comprehensive system-level testing. The system-level evaluations focused on three key tasks: restocking-only, retrieval-only, and full-cycle loop testing.


\begin{table*}[!ht]
\centering
\caption{Pipeline Performance Summary}
\label{pipeline_performance}
\begin{tabular}{lcccccc}
\toprule
\textbf{Pipeline} & \textbf{\# of Tests} & \textbf{\# of Successes} & \textbf{Avg Time} & \textbf{Recovery Triggered} & \textbf{Success Rate} \\
\midrule
Navigation (-R)  & 20 & 18 & 14 sec & - & 90.0\% \\
Navigation  & 20 & 19 & 16 sec & 1 & 95.0\% \\
Constrained Motion Planning (-R)  & 20 & 13 & 3 sec & - & 65.0\% \\
Constrained Motion Planning  & 20 & 18 & 5 sec & 8 & 90.0\% \\
Scene Understanding (-R)  & 20 & 13 & 67 sec & - & 85.0\% \\
Scene Understanding & 20 & 20 & 88 sec & 6 & 100.0\% \\
Suction Pose Generation  & 20 & 17 & \textless1 sec & - & 85.0\% \\
\bottomrule
\end{tabular}
\end{table*}

\subsection{Pipeline Performance}
Table~\ref{pipeline_performance} presents individual pipeline performance. We ablated the recovery mechanism from ORB, denoted as (-R), to evaluate its contribution to system robustness. Inference time includes any physical recovery actions, and “Recovery Triggered” indicates the count of such actions, if applicable.

For navigation tests, we teleoperated the robot to random locations after Nav2 localization. A trial was considered successful if the robot reached within 5 cm of the ground-truth goal. The variant without the ArUco-based recovery strategy (described in Section~\ref{nav}) is denoted as (–R).

For constrained motion planning, we positioned the manipulator near obstacles (e.g., shelves) with accurate collision representations. Targets consisted of physically feasible end-effector poses. Success followed the same criteria as free-space tests but required collision-free trajectories. The (-R) variant excluded recovery poses when direct path planning initially failed.

For scene understanding evaluations, the system identified and recorded the poses of all visible objects from a known robot viewpoint. The (-R) variant disabled recovery attempts that adjusted head angles to re-scan the scene if necessary.

For suction pose generation tests, we randomly placed inventory items within the robot’s view and workspace. The pipeline combined suction pose generation with free-space motion planning. Success required positioning the end effector within 1 cm of the object’s surface and maintaining orientation within 5 degrees of the surface normal. We did not explicitly evaluate the system's robustness to lighting condition due to the controlled indoor space assumption. Perception pipeline's collective inference time for object detection, segmentation, and suction pose generation consistently remained under 1 second despite the integration of computationally intensive models like SAM2 on onboard NVIDIA Jetson AGX Orin. This ensures the perception subsystem does not bottleneck the system, maintaining real-time responsiveness crucial for OR logistics tasks.

\subsection{Integrated Performance}
We conducted a complete system-level evaluation on the Fetch robot in a controlled environment setup. The results are reported in Table~\ref{sys_performance}. Each Retrieval Test follows the following procedure: A user requests an item from the shelf, the robot navigates to the shelf location, picks the object, and carries it back to the user's location. We determined success if the robot executed the sequences without human intervention, i.e., gave the user the item that he/she wanted.

Each Restocking Test follows the following procedure:  The robot observes the environment, identifies the object of interest, picks up the item, and drops the item inside the bin associated with that specific item. We determined success as the robot picking up the correct item and dropping the item in the correct bin. 

The combined test was used to simulate the actual workflow that the robot may encounter in an operating room supply room setting.
It was tasked with repeating one retrieval per four restocking actions. We determined success by observing that the robot performed these five tasks successfully without human intervention.

\begin{table}[h!]
\centering
\caption{System Performance Summary}
\label{sys_performance}
\begin{tabular}{lcccc}
\toprule
\textbf{Test Case} & \textbf{\# of Tests} & \textbf{\# of Succ.} & \textbf{Avg Time} & \textbf{Succ. Rate} \\
\midrule
Retrieval  & 20 & 16 & 3 min  & 80\%  \\
Restocking & 50 & 48 & 2 min  & 96\%  \\
Combined   & 12 & 11 & 12 min & 92\%  \\
\bottomrule
\end{tabular}
\vspace{-15pt}
\end{table}

\subsection{Failure Cases}
While ORB's architecture attempts to minimize potential failures, we observed occasional failures due to the following conditions. Each occurrence is given as a fraction of the total failure cases across all experiments. 
\begin{itemize}
    \item \textbf{Navigation (40 tests):} Failed to reach the goal within a tight tolerance (2/3); Open-loop control during fallback failed to drive robot within tolerance (1/3).
    \item \textbf{Motion Planning (80 tests)}: Start-pose beyond configured URDF; (5/12). Failed to register key voxels in the scene, resulting in a collision (1/12); Planner failed to find a solution (6/12).
    \item  \textbf{Scene Understanding (40 tests)}: ArUco tag misidentified (1/7); Missed ArUco tag during search (2/7); Open-vocabulary search (Grounded DINO) not finding object in scene (4/7).
    \item  \textbf{Suction Pose Generation (20 tests)}: Segmentation mask covering more than one object (1/3); Orientation is not within surface normal tolerance (2/3). 
\end{itemize}
We also provide failure cases for system-level tests:
\begin{itemize}
    \item \textbf{Retrieval:} Suction grasp failure due to object deformation (3/4); Failed to plan for manipulation (1/4);
    \item \textbf{Restocking:}  Suction grasp failure due to object deformation (1/2); Failed to generate proper grasp pose due to segmentation mask covering two objects (1/2);
    \item \textbf{Combined:} Failed to generate proper suction pose due to segmentation mask covering two objects (1/1);
\end{itemize}

\section{Conclusion}
We presented ORB, an autonomous mobile manipulator for logistics in OR environments. ORB integrates a robust software architecture, featuring modular behavior trees, advanced perception capabilities using state-of-the art models, and real-time execution on an Nvidia Jetson AGX Orin. Real-world experiments showed ORB completing operating room logistics tasks and failure recovery mechanisms improving reliability.

With pipeline performance exceeding 80\,\%, we demonstrate an effective system; however, several improvements can further enhance success rates. One approach is leveraging GPU parallel processing and cuRobo to evaluate the best $k$ grasp poses simultaneously, reducing failure-to-plan scenarios without sacrificing speed. Additionally, incorporating feedback from grasp attempts to train a learned suction pose generation model would enable continuous improvement on real-world. Finally, while our current segmentation method is reliable, it depends on fine-tuned bounding box generation. A general one-shot template matching approach~\cite{liu2024matchersegmentshotusing} could eliminate this need, streamlining the pipeline.

\FloatBarrier
\addtolength{\textheight}{-13cm}   









\bibliographystyle{IEEEtran}
\bibliography{ref.bib}

\begin{thebibliography}{10}
\providecommand{\url}[1]{#1}
\csname url@samestyle\endcsname
\providecommand{\newblock}{\relax}
\providecommand{\bibinfo}[2]{#2}
\providecommand{\BIBentrySTDinterwordspacing}{\spaceskip=0pt\relax}
\providecommand{\BIBentryALTinterwordstretchfactor}{4}
\providecommand{\BIBentryALTinterwordspacing}{\spaceskip=\fontdimen2\font plus
\BIBentryALTinterwordstretchfactor\fontdimen3\font minus \fontdimen4\font\relax}
\providecommand{\BIBforeignlanguage}[2]{{%
\expandafter\ifx\csname l@#1\endcsname\relax
\typeout{** WARNING: IEEEtran.bst: No hyphenation pattern has been}%
\typeout{** loaded for the language `#1'. Using the pattern for}%
\typeout{** the default language instead.}%
\else
\language=\csname l@#1\endcsname
\fi
#2}}
\providecommand{\BIBdecl}{\relax}
\BIBdecl

\bibitem{moxi_web}
\BIBentryALTinterwordspacing
D.~Robotics, ``Moxi: The hospital robot assistant,'' 2024, accessed: 2025-03-15. [Online]. Available: \url{https://www.diligentrobots.com/moxi}
\BIBentrySTDinterwordspacing

\bibitem{curobo}
B.~Sundaralingam, S.~K.~S. Hari, A.~Fishman, C.~Garrett, K.~V. Wyk, V.~Blukis, A.~Millane, H.~Oleynikova, A.~Handa, F.~Ramos, N.~Ratliff, and D.~Fox, ``curobo: Parallelized collision-free minimum-jerk robot motion generation,'' 2023.

\bibitem{thakar2022survey}
S.~Thakar, S.~Srinivasan, S.~Al-Hussaini, P.~Bhatt, P.~Rajendran, Y.~J. Yoon, N.~Dhanaraj, R.~Malhan, M.~Schmid, V.~Krovi, and S.~Gupta, ``A survey of wheeled mobile manipulation: A decision making perspective,'' \emph{Journal of Mechanisms and Robotics}, vol.~15, pp. 1--38, 05 2022.

\bibitem{technologies9010008}
M.~Kyrarini, F.~Lygerakis, A.~Rajavenkatanarayanan, C.~Sevastopoulos, H.~R. Nambiappan, K.~K. Chaitanya, A.~R. Babu, J.~Mathew, and F.~Makedon, ``A survey of robots in healthcare,'' \emph{Technologies}, vol.~9, no.~1, 2021.

\bibitem{tug}
R.~Bloss, ``Mobile hospital robots cure numerous logistic needs,'' \emph{Industrial Robot: An International Journal}, vol.~38, pp. 567--571, 10 2011.

\bibitem{Deepak03032016}
B.~Deepak and D.~R. Parhi, ``Control of an automated mobile manipulator using artificial immune system,'' \emph{Journal of Experimental \& Theoretical Artificial Intelligence}, vol.~28, no. 1-2, pp. 417--439, 2016.

\bibitem{health_mm}
A.~Kapusta, P.~Grice, H.~Clever, Y.~Chitalia, D.~Park, and C.~Kemp, ``A system for bedside assistance that integrates a robotic bed and a mobile manipulator,'' \emph{PLOS ONE}, vol.~14, p. e0221854, 10 2019.

\bibitem{506501}
U.~Nassal and R.~Junge, ``Fuzzy control for mobile manipulation,'' in \emph{Proceedings of IEEE International Conference on Robotics and Automation}, vol.~3, 1996, pp. 2264--2269 vol.3.

\bibitem{Colledanchise_2018}
M.~Colledanchise and P.~Ögren, ``Behavior trees in robotics and ai,'' Jul. 2018.

\bibitem{Zhao_2024}
Z.~Zhao, S.~Cheng, Y.~Ding, Z.~Zhou, S.~Zhang, D.~Xu, and Y.~Zhao, ``A survey of optimization-based task and motion planning: From classical to learning approaches,'' \emph{IEEE/ASME Transactions on Mechatronics}, p. 1–27, 2024.

\bibitem{black2024pi0visionlanguageactionflowmodel}
K.~Black, N.~Brown, D.~Driess, A.~Esmail, M.~Equi, C.~Finn, N.~Fusai, L.~Groom, K.~Hausman, B.~Ichter, S.~Jakubczak, T.~Jones, L.~Ke, S.~Levine, A.~Li-Bell, M.~Mothukuri, S.~Nair, K.~Pertsch, L.~X. Shi, J.~Tanner, Q.~Vuong, A.~Walling, H.~Wang, and U.~Zhilinsky, ``$\pi_0$: A vision-language-action flow model for general robot control,'' 2024.

\bibitem{Bavle_2023}
H.~Bavle, J.~L. Sanchez-Lopez, C.~Cimarelli, A.~Tourani, and H.~Voos, ``From slam to situational awareness: Challenges and survey,'' \emph{Sensors}, vol.~23, no.~10, p. 4849, May 2023.

\bibitem{Cao_2021}
H.~Cao, H.-S. Fang, W.~Liu, and C.~Lu, ``Suctionnet-1billion: A large-scale benchmark for suction grasping,'' \emph{IEEE Robotics and Automation Letters}, vol.~6, no.~4, p. 8718–8725, Oct. 2021.

\bibitem{mahler2019learning}
J.~Mahler, M.~Matl, V.~Satish, M.~Danielczuk, B.~DeRose, S.~McKinley, and K.~Goldberg, ``Learning ambidextrous robot grasping policies,'' \emph{Science Robotics}, vol.~4, no.~26, p. eaau4984, 2019.

\bibitem{orthey2023samplingbasedmotionplanningcomparative}
A.~Orthey, C.~Chamzas, and L.~E. Kavraki, ``Sampling-based motion planning: A comparative review,'' 2023.

\bibitem{wise_fetch_2016}
M.~Wise, M.~Ferguson, D.~King, M.~Diehr, and D.~Dymesich, ``Fetch \& freight: Standard platforms for service robot applications,'' in \emph{Workshop on Autonomous Mobile Service Robots}, 2016.

\bibitem{chengzhou_vacuum_gripper}
\BIBentryALTinterwordspacing
S.~C. Technology, ``Small size electric vacuum gripper (cz-evs01),'' 2025, accessed: 2025-03-15. [Online]. Available: \url{http://bit.ly/4hdmZlW}
\BIBentrySTDinterwordspacing

\bibitem{nvidia_jetson_orin}
\BIBentryALTinterwordspacing
N.~Corporation, ``Jetson orin series: Embedded ai computing platform,'' 2025, accessed: 2025-03-15. [Online]. Available: \url{https://www.nvidia.com/en-us/autonomous-machines/embedded-systems/jetson-orin/}
\BIBentrySTDinterwordspacing

\bibitem{BehaviorTreeCPP}
\BIBentryALTinterwordspacing
D.~Faconti, ``Behaviortree.cpp: The c++ library to create behavior trees,'' 2025, gitHub repository. [Online]. Available: \url{https://github.com/BehaviorTree/BehaviorTree.CPP}
\BIBentrySTDinterwordspacing

\bibitem{liu2024grounding}
S.~Liu, Z.~Zeng, T.~Ren, F.~Li, H.~Zhang, J.~Yang, Q.~Jiang, C.~Li, J.~Yang, H.~Su, J.~Zhu, and L.~Zhang, ``{Grounding DINO}: Marrying {DINO} with grounded pre-training for open-set object detection,'' in \emph{Proceedings of the European Conference on Computer Vision (ECCV)}, 2024.

\bibitem{GARRIDOJURADO20142280}
S.~Garrido-Jurado, R.~Muñoz-Salinas, F.~Madrid-Cuevas, and M.~Marín-Jiménez, ``Automatic generation and detection of highly reliable fiducial markers under occlusion,'' \emph{Pattern Recognition}, vol.~47, no.~6, pp. 2280--2292, 2014.

\bibitem{wang2022yolov7trainablebagoffreebiessets}
C.-Y. Wang, A.~Bochkovskiy, and H.-Y.~M. Liao, ``Yolov7: Trainable bag-of-freebies sets new state-of-the-art for real-time object detectors,'' 2022.

\bibitem{ravi2024sam2segmentimages}
N.~Ravi, V.~Gabeur, Y.-T. Hu, R.~Hu, C.~Ryali, T.~Ma, H.~Khedr, R.~Rädle, C.~Rolland, L.~Gustafson, E.~Mintun, J.~Pan, K.~V. Alwala, N.~Carion, C.-Y. Wu, R.~Girshick, P.~Dollár, and C.~Feichtenhofer, ``Sam 2: Segment anything in images and videos,'' 2024.

\bibitem{zhou2018open3dmodernlibrary3d}
Q.-Y. Zhou, J.~Park, and V.~Koltun, ``Open3d: A modern library for 3d data processing,'' 2018.

\bibitem{nvblox}
A.~Millane, H.~Oleynikova, E.~Wirbel, R.~Steiner, V.~Ramasamy, D.~Tingdahl, and R.~Siegwart, ``nvblox: Gpu-accelerated incremental signed distance field mapping,'' 2024.

\bibitem{macenski2020marathon2}
S.~Macenski, F.~Martin, R.~White, and J.~Ginés~Clavero, ``The marathon 2: A navigation system,'' in \emph{2020 IEEE/RSJ International Conference on Intelligent Robots and Systems (IROS)}, 2020.

\bibitem{liu2024matchersegmentshotusing}
Y.~Liu, M.~Zhu, H.~Li, H.~Chen, X.~Wang, and C.~Shen, ``Matcher: Segment anything with one shot using all-purpose feature matching,'' 2024.

\end{thebibliography}

\end{document}